\documentclass[oneside,ngerman]{article}

\usepackage[utf8]{inputenc}
\usepackage{graphicx}
\usepackage{multirow}
\usepackage{hyperref}
\usepackage{float}
\usepackage[T1]{fontenc}
\usepackage{amsmath,amssymb}
\usepackage{caption}
\usepackage{color}
\usepackage{wrapfig}
\usepackage[font=small]{subcaption}
\usepackage{float}
\usepackage{bbding}
\usepackage{multirow}
\let\vec\mathbf
\newcommand{\ds}{\displaystyle}

\newcommand{\vecp}{\vec{p}}
\newcommand{\mat}[1]{\text{\textbf{#1}}}

\title{\vspace{-2.5cm}{\bf Adaptive neural domain refinement for solving time-dependent differential equations}}
\author{Toni Schneidereit and Michael Breuß \\ 
\\[1ex]
\small Applied Mathematics Group \\ 
\small Brandenburg University of Technology Cottbus-Senftenberg \\
\small Platz der Deutschen Einheit 1, 03046 Cottbus, Germany \\
\small \{Toni.Schneidereit,breuss\}@b-tu.de
}

\date{\small \today} 

\begin{document}

\maketitle

\begin{abstract}
A classic approach for solving differential equations with 
neural networks builds upon neural forms, 
which employ 
the differential equation with a discretisation of the solution domain. 
Making use of neural forms for time-dependent differential equations,
one can apply the recently developed method of domain 
fragmentation. That is, the domain may be split into several
subdomains, on which the optimisation problem is solved. 
\par 
In classic adaptive numerical methods, 
the mesh as well as the domain may be
refined or decomposed, respectively, in order to improve accuracy. 
Also the degree of approximation accuracy may be adapted. 
It would be desirable to transfer such important and successful
strategies to the field of neural network based solutions.
In the present work, we propose a novel adaptive neural approach 
to meet this aim for solving time-dependent problems.
\par
To this end, each subdomain is reduced in size until the 
optimisation is resolved up to a predefined training accuracy. 
In addition, while the neural networks employed are by default 
small, we propose a means to adjust also the number of neurons 
in an adaptive way. We introduce conditions to automatically 
confirm the solution reliability and optimise computational parameters 
whenever it is necessary. Results are provided for several 
initial value problems that illustrate important 
computational properties of the method alongside.
In total, our approach not only allows to analyse in high detail
the relation between network error and numerical accuracy.
The new approach also allows reliable neural network solutions over 
large computational domains.
\end{abstract}
\begin{center}
{\small \textbf{\textit{Keywords:}} neural forms, differential equations, physics-informed neural networks, adaptive neural refinement, domain decomposition}
\vspace{0.875cm}
\end{center}

\section{Introduction}

Differential equations (DEs) are important models in many areas of science and engineering, as they often represent real-world 
phenomena \cite{Antia2012book}. A special class of DEs are initial value problems, describing the time evolution of a system. 
The variety of neural network approaches for solving DEs 
has increased over the last years and decades \cite{Maede1994lin,Yadav2015IntroNN,Dissanayake1994approx}. 
They mostly focus on obtaining a cost function out of the 
DE structure and given initial or boundary conditions. 
The cost function in this context has the characteristic of connecting the DE with the neural network framework \cite{Schneidereit2021SCNF,mall2016application}. 
This may be achieved with so-called neural forms (NFs), 
which are in fact trial solutions satisfying the given conditions \cite{lagaris1998artificial,Lagari2020Neural,Schneidereit2020ODEANN}.
The neural forms approach results in an unsupervised learning framework. In the end, the neural form represents the solution 
of the DE. \par
Other neural approaches combine unsupervised and supervised 
training, where the neural network outcome is compared to 
true (known) data corresponding to certain domain data \cite{piscopo2019solving,lagaris2000boundary,Raissi2019PINN,Jagtap2020cPINN}. 
Typically the unsupervised part arises from the DE structure, 
while given initial or boundary conditions are directly added to 
the cost function and are treated in a supervised way \cite{Blechschmidt2021Three}. The resulting difference, the error, 
is then used for learning the adjustable neural network parameters. 
Therefore, the neural network itself represents the solution 
of the DE after training in such approaches. 
\par 
Turning to classical numerical methods for solving all kinds 
of differential equations, one may consider e.g.\ 
Runge-Kutta methods \cite{Hairer1993ODE1,Hairer1996ODE2} 
for time integration or the finite element
method (FEM) \cite{Zienkiewicz2005FEM2}. In order to obtain high
accuracy and robustness, many numerical schemes feature adaptive
mechanisms regarding, e.g., step size control \cite{Antia2012book,Hairer1993ODE1} 
or mesh refinement
\cite{Bangerth2003AFEM,Berger1984AMR1,Verfuerth1994estimate}. 
That is, certain areas 
of the solution domain may require more elements or grid points, 
in other words a refined mesh, for improved reliability and 
accuracy. Such adaptive mesh refinement techniques enable the 
mesh to be locally refined based on a suitable error estimate.
\par 
Several works offer neural network based strategies and approaches 
to generate optimal meshes or mesh refinements for use with the 
finite element method \cite{Alfonzetti1998adaNN,Bohn2021RNNrefine}.
Predicting areas which are of interest in the sense of a 
required mesh refinement using neural networks is the 
objective of \cite{Manevitz2005TimeSeries}. 
Their time-series analysis is employed to predict 
elementwise solution gradient values. The used neural network 
yields an indicator based on local gradient values in 
space and time. This indicator is then used to predict whether 
a mesh refinement or a coarsening may be suitable. 
While in this method the mesh refinement indicator 
is realised by a neural network, the FEM is used for 
solving the PDE examples. Complementary to the latter 
approach, in \cite{Breuss2009flux} a learning strategy is 
developed which keeps the mesh fixed but selects 
the numerical scheme that gives locally high accuracy 
based on local gradients. 
\par
The most relevant related article in the context of our 
work may be the adaptive neural approach 
in \cite{Anitescu2019Second}, so let us discuss this work more in detail. It features a feedforward neural network in a 
framework combining both supervised and unsupervised terms, 
similar to \cite{piscopo2019solving,Raissi2019PINN,Jagtap2020cPINN}. 
The training process includes several evolution steps, 
each consisting of the optimisation over the training points 
combined with an evaluation of results at a finer grid. 
The latter is realised with the same set of neural network 
parameters obtained from the training step. It is proposed 
to start with a coarse grid and to perform local grid 
refinement whenever the resulting network errors differ. 
The method is developed for boundary value problems arising 
with stationary PDEs, like e.g.\ the Poisson equation. 
Results indicate that more complex neural network architectures
(w.r.t.\ number of layers and neurons) or more training points 
may increase the accuracy. 
\par
Let us stress that in the discussed work \cite{Anitescu2019Second},
the mesh is refined but treated in a global fashion and 
not decomposed into subdomains. However, we also find the 
combination of domain decomposition with neural 
networks \cite{Raissi2019PINN,Jagtap2020cPINN} which is also related 
to our method. The so-called physics-informed neural 
networks may be uniquely in predefined, discrete 
subdomains which may feature different network sizes. 
At the subdomain interfaces, a physical continuity 
condition holds and the average solution of the two 
corresponding neural networks at each interface is 
enforced. Several neural networks 
learn the governing DE locally instead of using a single 
one for the entire domain, which results overall 
in a small network error.
\par 
{\bf Problem statement and contribution.} In our previous work, we investigated the 
computational characteristics of a small feedforward neural network with only one hidden 
layer \cite{Schneidereit2020Study}. It is well-known that the parameters within
a neural network are not independent of each other. That is, changing one component of the 
setup may require to change other components to improve or at least to maintain the results. 
Computationally, larger domain sizes appear to be challenging for the neural forms approach 
\cite{lagaris1998artificial} together with the studied neural network setup. 
Turning to weight initialisation, the use of random initial weights allows to achieve results 
that can be considered as reliable, while they lead to variance in repeated computations, based on the initially generated values. The use of constant initial weights typically results in less accurate approximations, but using them leads to identical results in repeated computations
which may provide some advantages for an analysis. Based on these investigations, we proposed a collocation polynomial extension for the neural forms and a subdomain division approach, which splits the solution domain into equidistant subdomains \cite{Schneidereit2021SCNF}. Since the neural forms adopted from \cite{lagaris1998artificial} directly incorporate the initial condition in its construction, each temporal subdomain generates a new initial condition for the subsequent subdomain. As it turns out, both extensions to the original neural forms approach were able to improve the 
computational results with respect to weight initialisation and larger domain sizes.
However, equidistant subdomains may not be the optimal choice in regions where the solution 
is easy or difficult to learn. 

Therefore, we now propose the adaptive neural domain refinement (ANDRe), which makes use of the subdomain collocation (polynomial) neural forms (SCNF). These are optimised repeatedly over the domain. The domain itself is allowed to split into subdomains which may locally decrease in size, whenever the network error is not sufficiently small. Therefore, we combine the advantageous characteristics from domain decomposition and adaptive mesh refinement. Furthermore, we embed into the described process a means to adapt the number of neurons used for optimisation in each subdomain. 
This is done with the aim to increase reliability and accuracy of the approximation.

Thus we also combine adaptive refinement of the domain with adaptivity in the neural sense. 
In addition to that, the results open an opportunity to discuss the relation between neural network and numerical measurement metrics. 

The following section introduces the subdomain collocation neural form (SCNF), as well as the incorporated neural networks and the optimisation. Based on the SCNF, we continue to propose the adaptive domain refinement (ANDRe) algorithm. Later, this approach is applied to four initial value problems, each representing a different type. The results are discussed in detail and we finish the paper by a conclusion with an outlook to possible future work.

\section{The methods}

The overall aim is to solve initial value problems (IVPs) in form of 
\begin{equation}
G\big(t,u(t),\dot{u}(t)\big)=0,~~~u(t_0)=u_0,~~~t\in D\subset\mathbb{R}
\label{ivp-1}
\end{equation}
with given initial values $u(t_0)=u_0$. 
We identify $\ds \dot{u}(t)$ as the time derivative of $\ds u(t)$. 
Let us note at this point, that $\ds G$ may also denote a system of IVPs. 
In the following we will first focus on IVPs with only one equation and 
later provide the necessary information in order to extend the approach to 
systems of IVPs. \par 

\subsection{The subdomain collocation neural form (SCNF)} 
\label{sectionSCNF}

The neural forms approach \cite{lagaris1998artificial} seeks to replace the solution 
function with a differentiable trial solution
\begin{equation}
\tilde{u}(t,\vecp)=A(t)+F(t,\vecp)
\label{origNF}
\end{equation}  
which connects the given IVP with a neural network term, 
incorporating the weight vector $\ds \vecp$. 
In Eq.\ \eqref{origNF}, both $\ds A(t)$ and $F(t,\vecp)$ are problem specific and 
have to be constructed under the requirement of fulfilling the initial condition.
Besides replacing the solution function, its time derivative is expressed as well
by differentiating the constructed neural form.
\par
One of the possible neural form constructions, to which we will refer as classic neural form (NF) and which has been 
proposed in \cite{lagaris1998artificial}, is to set 
$\ds A(t)=u_0$ and $F(t,\vecp)=N(t,\vecp)(t-t_0)$. This configuration ensures to 
remove the impact of $\ds N(t,\vecp)$ at the initial point $\ds t_0$, whereas 
then $\ds \tilde{u}(t,\vecp)$ equals the initial value $\ds u_0$.
\par
The subdomain collocation neural form (SCNF) approach now opts to extend the classic NF (Eq.\ \eqref{origNF}) in two directions, 
{\em (i)} by increasing the polynomial degree of the term $\ds F(t,\vecp)$ 
as explained below and {\em (ii)} by introducing domain fragmentation, which is a novel 
approach to solve the initial value problem on subdomains in order to increase the numerical accuracy. 
\par 
Since $\ds u_0$ is a constant value, we find the resulting NF in Eq.\ \eqref{origNF} to resemble 
a first order polynomial in $\ds (t-t_0)$. Hence, we propose to extend the polynomial order of the classic neural form, inspired by collocation polynomials \cite{Antia2012book}. Therefore we transform
\begin{equation}
F(t,\vecp) \rightarrow F(t,\mat{P}_m) = \sum_{k=1}^m N_k(t,\vecp_k)(t-t_0)^k
\end{equation}
Here, $\ds m$ represents the SCNF order. This polynomial extension adds more flexibility to the approach. However, this is achieved in a 
different way than just increasing the number of hidden layer neurons in a single neural network, 
since additional networks arise that are connected to the factors $\ds (t-t_0)^k$, 
see \cite{Schneidereit2021SCNF}.
Thereby, the weight matrix $\ds \mat{P}_m$ stores each weight vector $\ds \vecp_k,k=1,\ldots,m$ 
as column vectors. \par
We thus discretise the domain $D$ by the collocation method employing a uniform grid 
with $n+1$ grid points $t_i$ ($t_0<t_1<\ldots<t_n$), so that our novel collocation neural forms approach 
for the IVP \eqref{ivp-1} leads to the $\ds \ell_2$ cost function formulation
\begin{equation}
E[\mat{P}_m]=\frac{1}{2(n+1)}\sum_{i=0}^n\bigg\{G\big(t_i,\tilde{u}_C(t_i,\mat{P}_m),\dot{\tilde{u}}_C(t_i,\mat{P}_m)\big)\bigg\}^2
\label{costFct1}
\end{equation}
\par
Since the domain variable in $(t_i-t_0)^k$ effectively acts as a scaling of $N_k(t_i,\vecp_k)$, we 
conjecture that a large domain size variation may introduce the need for a higher amount 
of training points or the use of a more complex neural network architecture.
Having this in mind, it appears very natural to couple the collocation neural forms with
a technique that refines the computational domain. To this end we will consider the non-adaptive version of domain fragmentation which opts to split the domain into separate a fixed number of equidistant subdomains.
\par
\begin{wrapfigure}[16]{r}{0.5\textwidth}
\includegraphics{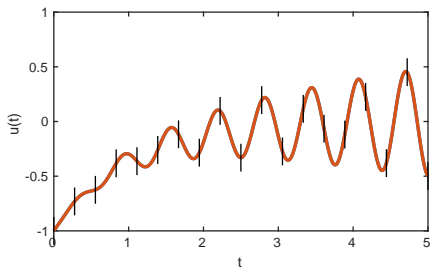}
\caption{\label{SCNFexample1} SCNF domain fragmentation example with fixed and equidistant subdomains, (orange) analytical IVP solution (cf. IVP in Eq.\ \eqref{IVPexample1}), (black/marked) subdomain boundaries, see \cite{Schneidereit2021SCNF} for details.}
\end{wrapfigure}
That said, we split the solution domain $\ds D$ into subdomains $\ds D_l,l=1,\ldots,h$, with $\ds n+1$ grid points $\ds t_{i,l}$ per domain fragment. Now the neural form is resolved separately in each subdomain. 
The interfacing grid points overlap, e.g., the computed value $\ds \tilde{u}_C(t_{n,l-1},\mat{P}_{m,l-1})$ 
at the last grid point of any subdomain $\ds D_{l-1}$ is set to be the new initial
value $\ds \tilde{u}_C(t_{0,l},\mat{P}_{m,l})$ for the next subdomain $\ds D_l$. The general idea is visualised in Fig.\ \ref{SCNFexample1}, where the black/vertical marks represent the equidistantly distributed subdomain boundaries for the solution of an example IVP.
\par 
Summarising the construction up to now, the SCNF satisfies the new 
initial values in each domain fragment, namely 
\begin{equation}
\tilde{u}_C(t_{i,l},\mat{P}_{m,l})=\tilde{u}_C(t_{0,l},\mat{P}_{m,l})+\sum_{k=1}^mN_k(t_{i,l},\vecp_{k,l})(t_{i,l}-t_{0,l})^k 
\label{SCNF}
\end{equation}
The neural networks are thus now scaled by $(t_{i,l}-t_{0,l})^k$, which in fact may avoid higher 
scaling factors as by the formulation over the entire domain, depending on the subdomain size. 
Incorporated in Eq.\ \eqref{costFct1}, the SCNF time derivative reads 
\begin{align}
\dot{\tilde{u}}_C(t_{i,l},\mat{P}_{m,l})=\sum_{k=1}^{m}\bigg[&\dot{N}_{k}(t_{i,l},\vecp_{k})(t_{i,l}-t_{0,l})^{k}\\
&+N_{k}(t_{i,l},\vecp_{k})k(t_{i,l}-t_{0,l})^{k-1}\bigg]\nonumber
\label{SCNFderiv}
\end{align}

In order to keep the overview of all terms and indices, we sum them up again: 
The $\ds i$-th grid point in the $\ds l$-th subdomain is denoted by $\ds t_{i,l}$, 
while $\ds t_{0,l}$ is the initial point in the subdomain $\ds D_l$ with the initial 
value $\ds \tilde{u}_C(t_{0,l},\mat{P}_{m,l})$. That is, $\ds t_{n,l-1}$ and $\ds t_{0,l}$
are overlapping grid points. In $\ds D_1$, $\ds \tilde{u}_C(t_{0,1},\mat{P}_{m,1})=u(t_0)$ holds. 
The matrix $\ds \mat{P}_{m,l}$ contains the set of the $m$ neural network weight vectors $\ds \vecp_k, k=1,\ldots,m$ in the 
corresponding subdomain. Finally, $\ds N_k(t_{i,l},\vecp_{k,l})$ denotes the $\ds k$-th neural 
network in $\ds D_l$. 
\par
The cost function employing the SCNF aims to minimise for each subdomain $\ds D_l$ the energy
\begin{equation}
E_l[\mat{P}_{m,l}]=\frac{1}{2(n+1)}\sum_{i=0}^n\bigg\{G\left(t_{i,l},\tilde{u}_C(t_{i,l},\mat{P}_{m,l}),\dot{\tilde{u}}_C(t_{i,l},\mat{P}_{m,l})\right)\bigg\}^2
\label{costFct2}
\end{equation}
Simultaneously to Eq.\ \eqref{costFct1}, the cost function now incorporates 
$\ds \tilde{u}_C(t_{i,l},\mat{P}_{m,l})$ instead of $\ds u(t)$. Let us mention at this point, that the $\ds (n+1)$ grid points in Eq.\ \eqref{costFct2} are used for the neural network training and refer to training points. While afterwards, the neural networks are used to verify the result with the learned weights and so called verification points. The latter are also grid points, but differently distributed than the training points. Therefore, the corresponding grid points will later be referred to as $\ds n_{_{TP}}$ with cost function notation $\ds E_l^{TP}[\mat{P}_{m,l}]$ and $\ds n_{_{VP}}$ with $\ds E_l^{VP}[\mat{P}_{m,l}]$, respectively.
\par 
If $\ds G$ in Eq.\ \eqref{ivp-1} represents a system of $\ds o$ IVPs, each solution function 
requires its own SCNF and the cost function derives from the sum over $\ds o$ 
separate $\ds \ell_2$-norm terms, i.e.\ one for each equation involved. 
We will address this extension in detail in the corresponding example later on.

\subsection{Neural network architecture and optimisation}

\begin{wrapfigure}[17]{r}{0.4\textwidth}
\includegraphics[scale=0.75]{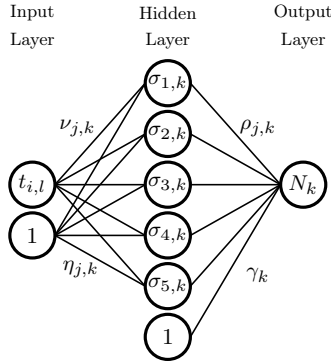}
\caption{\label{ANNarchitecture} Architecture of the $\ds k$-th neural network in the $\ds l$-th subdomain.}
\end{wrapfigure}

Let us start with an overview on the $\ds k$-th neural network architecture as displayed exemplary in Fig.\ \ref{ANNarchitecture}. 
The term $\ds N_k$, cf.\ Eq.\ \eqref{SCNF}, represents in general a feedforward neural network with one input layer neuron 
for the discretised domain data $\ds t_{i,l}$, $\ds H$ hidden layer neurons and one output 
layer neuron. In addition, both input layer and hidden layer incorporate one bias neuron. In general, the number of neurons in the hidden layer directly impacts the number of adjustable network weights, labeled as $\ds \nu_{j,k}$ (input layer neuron), $\ds \eta_{j,k}$ (input layer bias neuron), $\ds \rho_{j,k}$ (hidden layer neurons) and $\ds \gamma_k$ (hidden layer bias neuron) in Fig.\ \ref{ANNarchitecture}. These weights are stored in the weight vector $\ds \vecp_k$.
The neural network output reads 
\begin{equation}
N_k(t_{i,l},\vecp_k)=\sum_{j=1}^{H}\rho_{j,k}\sigma(z_{j,k})+\gamma_k
\label{NNout}
\end{equation}
Here, $\ds \sigma_{j,k}=\sigma(z_{j,k})=1/(1+e^{-z_{j,k}})$ represents the sigmoid activation function, with the weighted sum $\ds z_{j,k}=\nu_{j,k} t_{i,l}+\eta_{j,k}$. Therefore, $\ds H$ hidden layer neurons result in $\ds(3H+1)$ neural network weights in our framework.
The input layer passes the domain data $t_{i,l}$, weighted by $\nu_{j,k}$ and $\ds \eta_{j,k}$, to the hidden 
layer for processing. The neural network output $\ds N_k(t_{i,l},\vecp_k)$ is again a weighted sum of 
the values $\ds \rho_{j,k} \sigma(z_{j,k})$ and $\ds \gamma_k$. \par 
The neural network training is the process of minimising the cost function, cf.\ Eq.\ \eqref{costFct2}, with respect to the neural network weights $\ds \vecp_k$.
In addition, and in contrast to the $\ds (3H+1)$ weights for the $\ds k$-th neural network, the minimisation of $E_l[\mat{P}_{m,l}]$ requires $m(3H+1)$ weights to be adjusted, due to the SCNF order $m$. \par
In practice, the goal is to find a local minimum in the weight space (or energy landscape), which perhaps consists of many extreme points. This may be realised by first or second order optimisation techniques, which use the first or second cost function derivatives, respectively. One epoch of training includes the computation of the cost function gradient (first order derivatives) $\ds \nabla E_l[\mat{P}_{m,l}]$ w.r.t. the adjustable network weights, averaged over all training points (grid points). We will refer to this learning procedure as full batch training, since the neural network weights are updated only once per epoch. That is, the cost function and its gradient are computed and averaged with respect to all grid points in a subdomain. One subdomain may consist of, e.g., ten training points. Afterwards, the averaged (over all grid points of one subdomain) cost function gradient is used to update the weights. This usually takes several epochs for a successful training. 
In this paper, we use Adam optimisation \cite{kingma2017Adam} in order to update the neural network weights. With full batch training, the cost function also returns an (averaged) scalar value for each epoch. It provides information about the training status and whether the minimisation of $\ds E_l[\mat{P}_{m,l}]$ can be considered as accomplished or not. Let us recall, that each $\ds \vecp_k$ is separately optimised.  \par

\paragraph{Details on the optimisation} Let us consider the example IVP
\begin{equation}
\dot{u}(t)=t\sin(10t)-u(t),~~~u(0)=-1    
\end{equation}
for which we find $\ds G$, as a part of the cost function, in Eq.\ \eqref{costFct2} as
\begin{equation}
G=\dot{\tilde{u}}_C(t_{i,l},\mat{P}_{m,l})+\tilde{u}_C(t_{i,l},\mat{P}_{m,l})-t_{i,l}\sin(10t_{i,l})=0
\label{exampleIVP}
\end{equation}
The minimisation of Eq.\ \eqref{costFct2} aims to get $\ds G$ in Eq.\ \eqref{exampleIVP} as close to zero as possible. That is, the expression of interest actually reads
\begin{equation}
\dot{\tilde{u}}_C(t_{i,l},\mat{P}_{m,l})+\tilde{u}_C(t_{i,l},\mat{P}_{m,l})\approx t_{i,l}\sin(10t_{i,l})
\label{ApproxDurOpt}
\end{equation}
Here, the values of $\ds t_{i,l}\sin(10t_{i,l})$ are predetermined by the domain grid points, whereas the SCNF and its time derivative additionally depend on the neural network weights and their optimisation. Hence, Eq.\ \eqref{ApproxDurOpt} can be considered as satisfied for various combinations of $\ds \dot{\tilde{u}}_C(t_{i,l},\mat{P}_{m,l})+\tilde{u}_C(t_{i,l},\mat{P}_{m,l})$. Hence, the results may highly depend on the final location in the weight space. \par 
One reason for this circumstance may relate to the complexity of the energy landscape, which inherents multiple (local) minima that can lead to several combinations of the left hand side in Eq.\ \eqref{ApproxDurOpt}. Not all of these combinations must be real or useful solutions for the given domain training points. This issue may occur, e.g., when the initial weights are far away from a suitable minimum for a helpful approximation. When there is a minimum nearby the initialisation with unfavourable optimisation parameters, such that the optimiser can get stuck inside. However, fine tuning all the incorporated computational parameters is an ungrateful task since some of these are not independent of each others \cite{Schneidereit2020Study,Schneidereit2020ODEANN}. 

\section{The novel adaptive neural domain refinement (ANDRe)}

We propose in this section the embedding of the previously introduced SCNF approach into an adaptive algorithm. The resulting refinement strategy features two components, \textit{(i)} verification of the SCNF training status arising from the cost function (Eq.\ \eqref{costFct2}) in each subdomain serving as an error indicator and \textit{(ii)} an algorithmic component to perform the domain refinement.

\paragraph{\bf Algorithm summary.}

\begin{wrapfigure}[16]{r}{0.5\textwidth}
\includegraphics[scale=0.5]{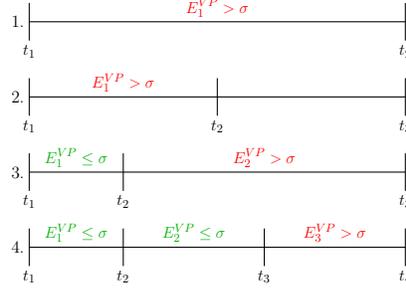}
\caption{\label{ANDReIdea} A visualisation of the basic idea behind ANDRe, with the error comparison and (sub-)domain split/reduction.}
\end{wrapfigure}

In Fig.\ \ref{ANDReIdea}, we consider an artificial example to sketch the principle behind ANDRe in a visual way. The basic idea is to optimise the cost function $\ds E_l[\mat{P}_{m,l}]$ for a given number of equidistant training points ($\ds n_{_{TP}}$) in each subdomain and to evaluate the results at equidistant verification points ($\ds n_{_{VP}}$), intermediate to $\ds n_{_{TP}}$. To obtain the subdomains, the algorithm starts with the cost function optimisation on the entire domain (Fig.\ \ref{ANDReIdea}(1.)). If the predefined verification error bound $\ds \sigma>0$ is not fulfilled, the domain is split in half. Now the optimisation task starts again for the left half since we only know the initial value for this subdomain. In case of the verification error $\ds E_l^{VP}[\mat{P}_{m,l}]$ (cost function evaluated for $\ds n_{_{VP}}$) again fails to go below $\ds \sigma$, the current (left) subdomain is reduced in size (see differences in Fig.\ \ref{ANDReIdea}(2.) to (3.)). Whereas a splitting is only performed when the computation takes place in the rightmost subdomain and $\ds \sigma$ is not satisfied by $\ds E_l^{VP}[\mat{P}_{m,l}]$, meaning
that the original right domain border is always kept and not shifted during refinement. The process of comparing the verification error to its error bound, reducing the current subdomain and starting the optimisation another time, is repeated until $\ds E_l^{VP}[\mat{P}_{m,l}]\le\sigma$. Therefore, in the artificial example in Fig.\ \ref{ANDReIdea}(3.), the leftmost subdomain is now considered to be learned. 
\par
Now, the process starts again for the rightmost subdomain (see Fig.\ \ref{ANDReIdea}(3.) and (4.)) with a new initial condition provided by the learned (left) subdomain. However, the current (new) subdomain starts at the right boundary of the first (learned) subdomain and ends at the right boundary of the entire domain. Therefore, the already learned subdomain is excluded from further computations. \par 
If a subdomain becomes too small or if the verification error increases after a subdomain split/reduction, the computational parameters are adjusted in a predefined, automated way. Details on the parameter adjustment will be provided in a corresponding paragraph later.
\par 
\medskip

\begin{figure}[!h]
\centering
\includegraphics{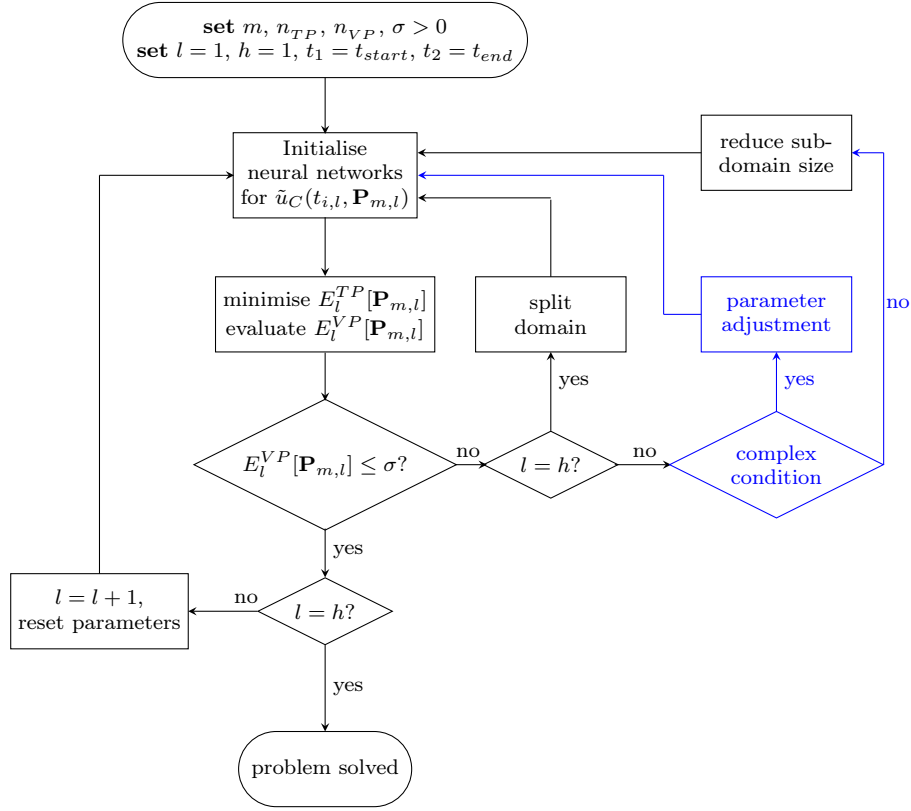}
\caption{\label{FlowChart1} Flowchart for the ANDRe algorithm.}
\end{figure}

Let us now provide detailed information about ANDRe, which is shown as a flowchart in Fig.\ \ref{FlowChart1}. Starting point is the choice of the SCNF order $\ds m$ and the subdomain resize parameter $\ds \delta$, which acts as a size reduction whenever a decrease is necessary. For optimisation we use equidistant training points $\ds n_{_{TP}}$. An important constant is the verification error bound $\ds \sigma>0$, used to verify the SCNF solution in the corresponding subdomain. After each complete optimisation, the results are evaluated by the cost function with the previous learned weights at intermediate verification points, resulting in $\ds E_l^{VP}[\mat{P}_{m,l}]$. The latter (scalar value) is then compared to $\ds \sigma$ in order to find out whether the solution can be considered as reliable or not.
We define $\ds l$ as the index of the subdomain, in which the SCNF is currently solved and $\ds h$ represents the total number of subdomains. The latter is not fixed and will increase throughout the algorithm. Finally, the very first domain is set to be the entire given domain $\ds D_1=[t_{start},t_{end}]=[t_1,t_2]$. Please note, while on the computational side, the subdomains are discretised and corresponding grid points denoted by $\ds t_{i,l}$, we only refer to subdomain boundaries by $\ds t_l$ in this paragraph, for simplicity. \par 

\paragraph{\bf Flowchart explanation.} The first processing operation 
\begin{equation}
\text{Initialise neural networks for}~\tilde{u}_C(t_{i,l},\mat{P}_{m,l})
\label{operation1}
\end{equation}
covers setting the initial architecture parameters such as number of hidden layer neurons, Adam learning rate and initialising the weights for $\ds \mat{P}_{m,l}$. \par  
Afterwards the optimisation problem 
\begin{align}
&\text{minimise}~E_l^{TP}[\mat{P}_{m,l}]\\
&\text{evaluate}~E_l^{VP}[\mat{P}_{m,l}]
\label{operation2}
\end{align}
is solved by training the SCNF framework for given equidistant $\ds n_{_{TP}}$ over the entire domain $D_1=[t_{1},t_{2}]$ (cf. Fig.\ \ref{ANDReIdea}(1.)). The evaluation for equidistant and intermediate $\ds n_{_{VP}}$ leads to the verification error $E_l^{VP}[\mat{P}_{m,l}]$.
\par 
Then the first decision block compares the verification error (after the training process has ended) to the error bound $\ds \sigma$:
\begin{equation}
E_l^{VP}[\mat{P}_{m,l}]\le\sigma?
\label{decision1}
\end{equation}
\begin{itemize}
\item[$\bullet$] \textbf{Eq. \eqref{decision1} NO}: In case the verification error did not go below $\ds \sigma$, the size of the current subdomain will be reduced. But first, another decision has to be made here. Namely, has $\ds E_l[\mat{P}_{m,l}]$ been solved for the first time on the current, rightmost (sub-)domain or in other words, is the current domain index $\ds l$ equal to number of total subdomains $\ds h$:
\begin{equation}
l=h?
\label{decision11}
\end{equation}
\begin{itemize}
\item[$\bullet$] \textbf{Eq. \eqref{decision11} YES}: That means the right boundary is $\ds t_{end}$ and we have to split the current subdomain $\ds l$ first, which leads to an increase of the number of total subdomains by $\ds 1$ ($\ds h=h+1$). The boundaries now have to be adjusted with the left one $\ds t_l$ to remain unchanged, while the former right boundary is now scaled by $\ds t_{l+1}=t_l+\delta(t_{l+1}-t_l)$, after $\ds t_{l+2}=t_{l+1}$ is set to be the right boundary of domain $\ds l+1$. For example, if an entire domain $\ds D_1=[t_{1},t_{2}]=[0,10]$ has to be split for the first time with $\ds \delta=0.5$, the resulting subdomains are $\ds D_1=[t_{1},t_{2}]=[0,5]$ and $\ds D_2=[t_{2},t_{3}]=[5,10]$. Afterwards, the algorithm leads back to Eq.\ \eqref{operation1}. 
\item[$\bullet$] \textbf{Eq. \eqref{decision11} NO}: In this case the current subdomain has already been split up. Now the right boundary has to be adjusted in order to decrease the current subdomain size. But beforehand we check for a complex condition (highlighted in blue) to ensure that a subdomain does not become too small. Additionally we also check if the verification error decreased compared to the prior computation on the same subdomain $\ds l$. That is, the algorithm compares the verification error from the formerly larger subdomain $\ds l$ to the current, size reduced subdomain $\ds l$. The condition itself may come in different shapes. We decided to check for one of the  
\begin{equation}
\begin{array}{c}
\text{complex conditions:} \\[1ex]
t_{l+1}-t_l\le 0.1?\\ 
\text{or}\\
E_l^{VP}~\text{from previous}~(l\ne h)~\text{subdomain}~\le~\text{current}~E_l^{VP}? 
\end{array}
\label{decision111}
\end{equation}
\begin{itemize}
\item[$\bullet$] \textbf{Eq. \eqref{decision111} YES}: At this point we employ a
\begin{equation}
\text{parameter adjustment}
\label{parameterAdjustment}
\end{equation} 
which may be realised problem specific and is later addressed in a corresponding paragraph. Afterwards, the algorithm leads back to Eq.\ \eqref{operation1}. Basically speaking, the adjustable parameters may include the number of hidden layer neurons, the learning rate, the number of training points and so on.
\item[$\bullet$] \textbf{Eq. \eqref{decision111} NO}: In this case, the subdomain is still large enough to be reduced in size while the verification error decays. Therefore we resize the right subdomain boundary $\ds t_{l+1}$ to
\begin{equation}
t_{l+1}=t_l+\delta(t_{l+1}-t_l)
\label{adjustBoundary}
\end{equation}
where $\ds \delta$ denotes the domain resize parameter. Continuing the example from above, resizing $\ds D_1$ of the already split domain leads to $\ds D_1=[t_{1},t_{2}]=[0,2.5]$ and $\ds D_2=[t_{2},t_{3}]=[2.5,10]$. Afterwards, the algorithm leads back to Eq.\ \eqref{operation1}. 
\end{itemize}
\end{itemize}
\item[$\bullet$] \textbf{Eq. \eqref{decision1} YES}: In case of the verification error being smaller or equal compared to $\ds \sigma$, the current subdomain $\ds l$ has been successfully learned by means of a sufficiently small verification error. Now it is necessary to determine, whether we are in the last subdomain (right boundary is $\ds t_{end}$) or if there is still one subdomain to solve the optimisation problem on, namely
\begin{equation}
l=h?
\label{decision12}
\end{equation}
\begin{itemize}
\item[$\bullet$] \textbf{Eq. \eqref{decision12} NO}: There is at least one subdomain left and therefore the current subdomain index is updated to $\ds l=l+1$ in order to solve the optimisation problem on the adjacent subdomain. Additionally we reset all the possibly adjusted parameters to the initial ones. Thus we make sure to not overuse the variable parameters in regions where the solution computes by using the initial ones. The algorithm then leads back to Eq.\ \eqref{operation1}.
\item[$\bullet$] \textbf{Eq. \eqref{decision12} YES}: All subdomains have been successfully learned and the initial value problem is entirely solved. 
\end{itemize}
\end{itemize}

We developed ANDRe in four steps, making it an adaptive neural algorithm for domain refinement. Excluding the blue part in Fig.\ \ref{FlowChart1}, the black part represents a fully functional algorithm that can refine the domain in an adaptive way with the focus laying on the verification error. Prior to this final version, the training error was used as the main training status indicator. The evaluation stage (verification error) on the other hand was later added, in addition to the training error. It turned out that small training errors do not necessarily result in a comparable numerical error, presumably due to possible overfitting. Therefore we included the verification stage, to reduce the impact of overfitting on the end result. However, we later recognised that the verification error has a much stronger relation to the numerical error. Therefore we were able to reduce the complexity by laying the focus directly on the verification. Furthermore, in some examples we recognised that the preset neural network architecture may not be flexible enough to learn certain subdomains. Hence, we upgraded ANDRe to incorporate an automated parameter adjustment mechanism, highlighted in Fig.\ \ref{FlowChart1} as blue. Whenever a subdomain becomes too small or the verification error in a subdomain increases compared the previous optimisation on the same subdomain (e.g., prior to a size reduction), network and optimisation related parameters may be rebalanced in a predefined way. We will later provide experimental evidence to prove the capabilities of ANDRe.

\section{Computational results and discussion}
\label{expSCNF}

In this section we discuss the computational results for different initial value problems (IVPs), solved by ANDRe. Beforehand, the framework parameters and methods are further specified. \par 

\paragraph{\bf Details on parameters and measurement metrics.} The neural forms approach comes with plenty parameters. We have already shown in a computational study \cite{Schneidereit2020Study}, that they are not independent of each other. Changing one parameter may require another parameter to be changed as well in order to improve or maintain the reliability. \par
\begin{figure}[!h]
\centering
\includegraphics{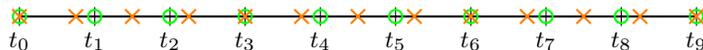}
\caption{\label{grid_distribution} Visualisation of grid point distribution in a subdomain, (green/circle) 10 equidistant training points $\ds t_i$, (orange/cross) 12 equidistant verification points. Please note that, e.g., $\ds n_{_{TP}}=9$ refers to ten training points in Tab.\ \ref{TabFixedParameters}.}
\end{figure}

 
\begin{wraptable}[15]{r}{0.5\textwidth}  
\caption{\label{TabFixedParameters} Initial computational parameters, ($^\ast$) part of parameter adjustment.} 
\begin{tabular}{l|l}
comp. parameter & value  \\
\hline
hidden layer neurons$\ds ^\ast$ & 5  \\
initial weight values & 0  \\
Adam learning rate$\ds ^\ast$ & 1e-3 \\
number of epochs & 1e5 \\
training points ($\ds n_{_{TP}}$) & 9 \\
verification points ($\ds n_{_{VP}}$) & 11 \\
SCNF order (m) & 5 \\
resize parameter ($\delta$) & 0.5 
\end{tabular}
\end{wraptable}
Tab.\ \ref{TabFixedParameters} lists the computational parameters which are initially fixed in our computational setup. Parameter marked with ($\ds ^\ast$) will be separately discussed in the corresponding paragraph. The initial weight values, the SCNF order as well as the number of epochs and training points ($\ds n_{_{TP}}$) have been previously investigated and are fixed to suitable values, see \cite{Schneidereit2020Study,Schneidereit2021SCNF} for further details. Nonetheless, each parameter has its impact on the solution. Key in training the neural networks are the training points $\ds t_i,~i=0,\ldots,9$, schematically depicted in Fig.\ \ref{grid_distribution} as green circles.  Generally speaking, Fig.\ \ref{grid_distribution} shows an arbitrary subdomain and the notation $\ds t_i$ was chosen for simplicity. From now on, the grid points in subdomain $\ds l$ are again referred to as $\ds t_{i,l}$ and follow the structure in Fig.\ \ref{grid_distribution}.

They serve as the input data and are important for the optimisation of the cost function
\begin{equation}
E_l^{TP}[\mat{P}_{m,l}]=\frac{1}{2(n+1)}\sum_{i=0}^{n_{_{TP}}}\bigg\{G\left(t_{i,l},\tilde{u}_C(t_{i,l},\mat{P}_{m,l}),\dot{\tilde{u}}_C(t_{i,l},\mat{P}_{m,l})\right)\bigg\}^2
\label{TPcost}
\end{equation}
in the $\ds l$-th subdomain. Let us comment on an optimisation procedure in some detail, referred to as incremental learning, employed in \cite{piscopo2019solving}. Here, the computation includes several complete optimisations per (temporarily untouched) subdomain. That is, for example with five increments in Fig.\ \ref{grid_distribution}, the training points are split into five sets and the first optimisation only takes $\ds t_0$ and $\ds t_1$ into account. Then, the second one uses $\ds t_0,t_1,t_2,t_3$ with the same weights from the first (complete) optimisation. This is continued until the optimisation uses all training points. The incremental learning procedure only applies to the training process and $\ds E_l^{TP}[\mat{P}_{m,l}]$, not to the verification. 

Speaking of that, the verification is performed with the cost function and the corresponding verification points ($\ds n_{_{VP}}$), which are differently distributed (cf. Fig.\ \ref{grid_distribution}) than the training points. With these discrete points and after the training process, the cost function returns a scalar value named verification error 
\begin{equation}
E_l^{VP}[\mat{P}_{m,l}]=\frac{1}{2(n+1)}\sum_{i=0}^{n_{_{VP}}}\bigg\{G\left(t_{i,l},\tilde{u}_C(t_{i,l},\mat{P}_{m,l}),\dot{\tilde{u}}_C(t_{i,l},\mat{P}_{m,l})\right)\bigg\}^2
\label{VPcost}
\end{equation}
As the naming suggests, this verification error is used to evaluate and verify the training results to indicate whether the IVP has been solved sufficiently well or not. 
For this purpose, the verification error bound $\ds \sigma$ will compare to Eq.\ \eqref{VPcost}. 

The domain resize parameter has also been fixed for all computations to $\ds \delta=0.5$. 
A larger value, up to $\delta=0.9$, would find individual subdomains faster due to the bigger size reduction but perhaps result in too many subdomains. On the other hand, a smaller value, down to $\delta=0.1$ may find the individual subdomains more carefully but would also heavily increase the computation time. We will discuss an experiment regarding the domain resize parameter later.

Tab.\ \ref{TabFixedParameters} will later also be extended by problem specific parameters, which are 
{\em (i)} verification error bound $\ds \sigma$, 
{\em (ii)} computational domain size, 
{\em (iii)} initial conditions and 
{\em (iv)} learning increments. These parameters will be specified and discussed in a subsequent paragraph. 

Turning to the measurement metrics for the results, we will compare ANDRe to the analytical solutions of four different initial value problems. We make use of the absolute value differences between the analytical solution and ANDRe in context of the (averaged) $\ds \ell_1$-norm $\Delta u_{l,1}$ and the $\ds \ell_{\infty}$-norm $\ds \Delta u_{l,\infty}$
\begin{align}
\Delta u_{l,1}&=\frac{1}{n+1}\sum_{i=0}^n\big|u(t_{i,l})-\tilde{u}_C(t_{i,l},\mat{P}_{m,l})\big| \\
\Delta u_{l,\infty}&=\max_{i}\big|u(t_{i,l})-\tilde{u}_C(t_{i,l},\mat{P}_{m,l})\big|
\end{align}
whereas $\Delta u_1$ and $\ds \Delta u_{\infty}$ average the numerical error over all subdomains
\begin{align}
\Delta u_1&=\frac{1}{h}\sum_{l=1}^h\Delta u_{l,1} \\
\Delta u_{\infty}&=\frac{1}{h}\sum_{l=1}^h\Delta u_{l,\infty}
\end{align}
The $\ell_{\infty}$-norm  basically returns the largest numerical error value. We will later refer to the corresponding norms as $\ds \ell_1$-error and $\ds \ell_{\infty}$-error. \par

\paragraph{\bf Details on parameter adjustment.} Let us now comment on the parameter adjustment since this part of the algorithm required a lot of fine tuning. After several experiments with different parameter adjustment methods, not documented here, the Adam learning rate and the number of hidden layer neurons were chosen to be a part of the parameter adjustment and may change during the process. Not only determine the hidden layer neurons the amount of adjustable weights, they are also connected to the universal approximation theorem \cite{Cybenko1989uat}. It basically states, that one hidden layer with a finite number of sigmoidal neurons is able to approximate every continuous function on a subset of $\mathbb{R}$. Since the finite number is not known beforehand, making the number of hidden layer neurons an adjustable parameter in this approach, seems reasonable and so does starting with a small amount (five neurons). \par 
The initial learning rate of Adam optimisation impacts how vast the location in the weight space changes after a weight update. Figuratively speaking, the larger the initial learning rate, the farther the optimiser can travel in the weight space, adding more flexibility and increasing the chance to find a suitable minimum. In this context, such a suitable minimum can be located at different positions, depending on the subdomain. It is not guaranteed by any means to find one near by the starting point. That motivates to start the computation with a fairly small initial learning rate (values taken from \cite{kingma2017Adam}) and to enable ANDRe to increase this value outside the optimisation cycle. That is, the initial learning rate can increase several times before the number of hidden layer neurons rearranges by two additional neurons. Adjusting the number of neurons resets the learning rate to its default parameter. \par
Has a subdomain in this way been successfully learned, both parameters are reset to their initial values. Let us recall, that the parameter adjustment does not take place during an optimisation cycle, it rather appears outside. In other words, we do not perturb the neural network training during the optimisation process. \par

\subsection{The evaluation of ANDRe for different initial value problems.} In \cite{Schneidereit2021SCNF} we have shown, that the SCNF with a fixed number of subdomains is capable of solving IVPs on larger domains. Increasing this number resulted in a decreasing numerical error. Now with ANDRe, we show that by demanding the network error to become sufficiently small in each subdomain, the algorithm can automatically determine a suitable number (and distribution) of the subdomains. \par
The following paragraph will introduce initial value problems (IVPs) for our evaluation. We have chosen these examples because \textit{(i)} each one represents a different IVP type, \textit{(ii)} expect for the last (system of IVPs) example, the analytical solutions are available and \textit{(iii)} each of them incorporates at least one interesting behaviour. However, the difficulty is limited because of \textit{(ii)}, but the focus of this paper does not lay on competitiveness in the first place. We rather show that the neural forms approach \cite{lagaris1998artificial,Schneidereit2020Study} benefits from our extension in terms of accuracy on large domains. In addition, this paper serves as an investigation of the relation between both numerical and neural network errors.

\paragraph{\bf Example IVPs and their analytical solutions.}

As a first example, we take on the following IVP with constant coefficients 
\begin{equation}
\left\{
\begin{array}{l}
\ds \dot{\psi}(t)-t\sin(10t)+\psi(t)=0,~~~\psi(0)=-1 \\[1em]
\ds \psi(t)=\sin(10t)\bigg(\frac{99}{10201}+\frac{t}{101}\bigg)+\cos(10t)\bigg(\frac{20}{10201}-\frac{10t}{101}\bigg)-\frac{10221}{10201}e^{-t}
\end{array}\right.
\label{IVPexample1} 
\end{equation}
which incorporates heavily oscillating and increasing characteristics, similar to instabilities. This example is still relatively simple and serves to demonstrate the main properties of our approach. We then proceed to an IVP with non-constant coefficients, that includes trigonometric and exponentially increasing terms:
\begin{equation}
\hspace*{-3cm}
\left\{
\begin{array}{l}
\ds \dot{\phi}(t)+\frac{\ds 1+\frac{1}{1000}e^t\cos(t)}{1+t^2}+\frac{2t}{1+t^2}\phi(t)=0,~~~\phi(0)=5 \\[1em]
\ds \phi(t)=\frac{1}{1+t^2}\bigg(-t-\frac{e^t\cos(t)}{2000}-\frac{e^t\sin(t)}{2000}+\frac{10001}{2000}\bigg)
\end{array}\right.
\label{IVPexample2}
\end{equation}
Furthermore, we choose to investigate the results for the non-linear IVP
\begin{equation}
\hspace*{-5cm}
\left\{
\begin{array}{l}
\ds \frac{\dot{\omega}(t)}{\cos^2(\omega(t))}\frac{1}{\cos^2(2t)}-2=0,~~~\omega(0)=\frac{\pi}{4} \\[1em]
\ds \omega(t)=\arctan\bigg(\frac{1}{4}\sin(4t)+t+1\bigg)
\end{array}\right.
\label{IVPexample3}
\end{equation}
which also has non-constant coefficients. Finally, we used ANDRe to solve the following non-linear system of IVPs
\begin{equation}
\hspace*{-5cm}
\left\{
\begin{array}{l}
\ds \dot{\tau}(t)=A\tau(t)-B\tau(t)\kappa(t),~~~\tau(0)=\tau_0 \\[1em]
\ds \dot{\kappa}(t)=-C\kappa(t)+D\tau(t)\kappa(t),~~~\kappa(0)=\kappa_0
\end{array}\right.
\label{IVPexample4}
\end{equation}
which is also known as the Lotka-Volterra equations \cite{Anisiu2014LV}, with parameters $A=1.5$, $B=1$, $C=3$, $D=1$. The initial values are $\ds \tau_0=3$, $\ds \kappa_0=1$, $\ds \kappa_0=3$, $\ds \kappa_0=5$ depending on the subsequent experiment. The chosen value for $\ds \kappa_0$ will be explicitly addressed. Since the Lotka-Volterra equations in Eq.\ \eqref{IVPexample4} do not have an analytical solution, we will compare the results to a numerical solution method, namely Runge-Kutta 4. \par 
We take the coupled IVPs in Eq.\ \eqref{IVPexample4} to demonstrate how the cost function for the neural forms approach reads. It is obtained as the sum of $\ds \ell_2$-norms of each equation, cf. Eq.\ \eqref{costFct2}. We use $\ds \tilde{\tau}_C=\tilde{\tau}_C(t_{i,l},\mat{P}_{m,l})$ and $\ds \tilde{\kappa}_C=\tilde{\kappa}_C(t_{i,l},\mat{P}_{m,l})$ as shortcuts:
\begin{equation}
\begin{array}{cc}
\ds E_l[\mat{P}_{m,l}]=\frac{1}{2(n+1)}\sum_{i=0}^n\bigg[&\ds\bigg\{\dot{\tilde{\tau}}_C-A\tilde{\tau}_C+B\tilde{\tau}_C\tilde{\kappa}_C\bigg\}^2+\\
\ds &\ds \bigg\{ \dot{\tilde{\kappa}}_C+C\tilde{\kappa}_C-D\tilde{\tau}_C\tilde{\kappa}_C\bigg\}^2\bigg]
\end{array}
\end{equation}
This equation is then subject to optimisation/training and verification.

\paragraph{\bf ANDRe and the analytical solutions.}

In this paragraph we demonstrate the results for applying ANDRe to the previously introduced example IVPs. We discuss the contrast to the analytical solutions and in case of Lotka-Volterra, to the numerical results provided by Runge-Kutta 4. In addition to the already given computational parameters in Tab.\ \ref{TabFixedParameters}, the problem specific parameters are listed in Tab.\ \ref{TabProbSpecParameters}. The corresponding initial conditions are given with the examples above. 

\begin{wraptable}[15]{r}{0.55\textwidth}  
\caption{\label{TabProbSpecParameters} Problem specific parameters, ($\ds \sigma$) represents the verification error bound, (inc) is short for increments and refers to the learning procedure discussed in context of Fig.\ \ref{grid_distribution}.}
\begin{tabular}{c|c|c|c}
Example & domain & $\ds \sigma$  & inc. \\
\hline
IVP in Eq.\ \eqref{IVPexample1} & $\ds t\in[0,15]$ & $\ds 1$e-5 & $\ds 5$ \\
IVP in Eq.\ \eqref{IVPexample2} & $\ds t\in[0,25]$ & $\ds 1$e-4 & $\ds 5$ \\
IVP in Eq.\ \eqref{IVPexample3} & $\ds t\in[0,20]$ & $\ds 1$e0 & $\ds 2$ \\
IVP in Eq.\ \eqref{IVPexample4} & $\ds t\in[0,30]$ & $\ds 1$e-3 & $\ds 5$ \\
\end{tabular}
\end{wraptable}

The domain sizes are chosen in this way, so that interesting parts in the analytical solution are visible and as challenges available for ANDRe. During the experimental testing, we recognised that the neural network errors and especially the verification error $\ds E_l^{VP}[\mat{P}_{m,l}]$ were not becoming arbitrarily small. In addition, the experiments revealed the problem specific dependencies of (local) minima locations in the weight space. Therefore we had to find (in an experimental way) the verification error bounds $\ds \sigma$ for each example IVP.   

\begin{table}[!h] 
\centering
\caption{\label{TabNumResults} Overview of the numerical results for the example IVPs in Eq.\ \eqref{IVPexample1}--Eq.\ \eqref{IVPexample4}, (h) total number of learned subdomains.}
\scalebox{1.25}{%
\begin{tabular}{c|c|c|c|c}
Example & domain & $\ds h$ & $\ds \ell_{1}$-error & $\ds \ell_{\infty}$-error \\ 
\hline
IVP in Eq.\ \eqref{IVPexample1} & $\ds t\in[0,15]$ & 113 & 1.4499e-4 & 1.9268e-4\\
IVP in Eq.\ \eqref{IVPexample2} & $\ds t\in[0,25]$ & 50 & 6.8152e-4 & 9.8980e-4\\
IVP in Eq.\ \eqref{IVPexample3} & $\ds t\in[0,20]$ & 32 & 4.6545e-3 & 4.9861e-3\\
IVP in Eq.\ \eqref{IVPexample4} & $\ds t\in[0,30]$ & 51 & - & - \\
\end{tabular}}
\end{table}

\begin{figure}[!h]
  \centering
  \subcaptionbox{\label{FIGIVPexample1a} entire domain}{
  \includegraphics[scale=1]{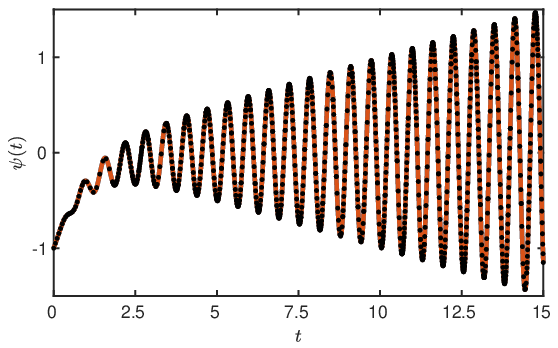}}
  \subcaptionbox{\label{FIGIVPexample1b} clipped domain}{
  \includegraphics[scale=1]{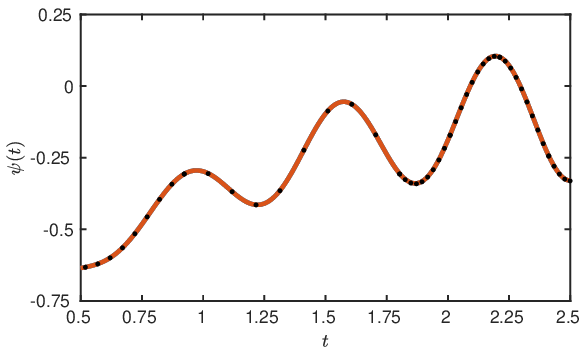}}
\caption{\label{FIGIVPexample1} {\bf IVP in Eq.\ \eqref{IVPexample1}} Comparison between (orange/solid) analytical solution and (black/dotted) ANDRe solution.}
\end{figure}

In Fig.\ \ref{FIGIVPexample1}, both the analytical solution (orange/solid) and ANDRe solution (black/dotted) are shown for the IVP in Eq.\ \eqref{IVPexample1}. Tab.\ \ref{TabNumResults} shows that 113 subdomains were necessary in order to satisfy the chosen verification error bound. The corresponding (averaged) $\ds \ell_1$-error indicates a decent behaviour, which we consider to represent a reliable solution to the IVP. It also compares to the results from our SCNF experiments \cite{Schneidereit2021SCNF} (predefined equidistant subdomain distribution). The same IVP, solved with 100 equidistant subdomains, returned an $\ds \ell_1$-error of 1.4339e-4. Therefore, ANDRe maintains the solution accuracy and comes with an advanced measurement metric. \par 
\begin{wrapfigure}[17]{r}{0.49\textwidth}
  \includegraphics{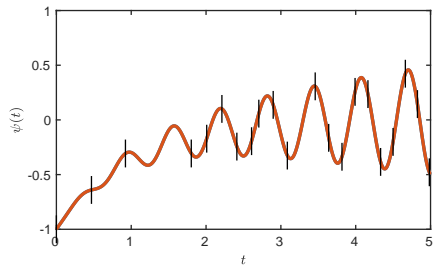}  
  \caption{\label{SCNFexample2} {\bf IVP in Eq.\ \eqref{IVPexample1}} ANDRe subdomain distribution for a cut-out of Fig.\ \ref{FIGIVPexample1a}, (orange/solid) analytical solution, (black/marked) subdomain boundaries, cf.\ Fig.\ \ref{SCNFexample1}.}
\end{wrapfigure}
The total number of training points for all subdomains is not equidistantly distributed. This circumstance is demonstrated in Fig.\ \ref{FIGIVPexample1b} for a clipped domain of Fig.\ \ref{FIGIVPexample1a}. Because of the general trend of the solution, we expect the density to be higher at the peaks and dips, while declining in between. However, the subdomain $\ds D_3=[0.9229,1.8027]$ is fairly large and includes two peaks and almost two dips as well. Compared to its adjacent subdomain $\ds D_4=[1.8027,2.0089]$, the size of $\ds D_3$ is unique, but also has a lower numerical error assigned. So the (local) numerical error in one subdomain, as well as the subdomain size itself do not necessarily share the global behaviour, where a higher amount of subdomains leads to a decreasing numerical error. \cite{Schneidereit2021SCNF} \par
Fig.\ \ref{SCNFexample2} shows the subdomain distribution related to Fig.\ \ref{FIGIVPexample1} in the beginning for $\ds D=[0,5]$. We find the domain size adjustment parameter $\ds \delta$ to show a significant influence here. It appears to be very important where one subdomain ends, because this may cause the adjacent one to be more difficult to solve. Please note that this statement holds under the consideration of the chosen neural network parameters. \par 

\begin{figure}[!h]
  \centering
  \subcaptionbox{\label{FIGIVPexample2a} entire domain}{
  \includegraphics[scale=0.95]{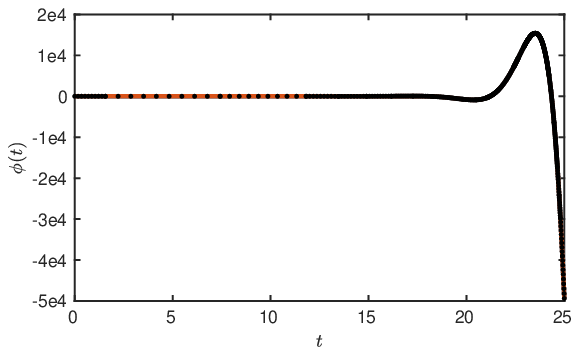}}
  \quad
  \subcaptionbox{\label{FIGIVPexample2b} clipped domain}{
  \includegraphics[scale=0.95]{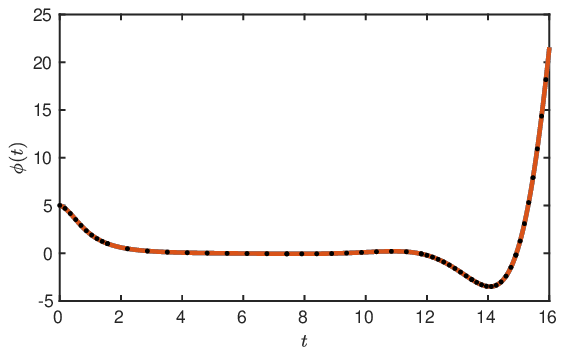}}
  \\
  \subcaptionbox{\label{FIGIVPexample2c} clipped domain}{
  \includegraphics[scale=0.95]{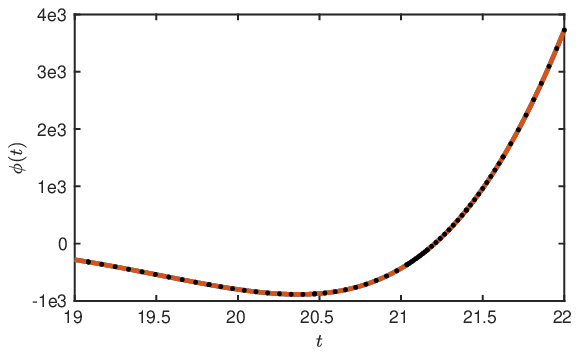}}
  \quad
  \subcaptionbox{\label{FIGIVPexample2d} clipped domain}{
  \includegraphics[scale=0.95]{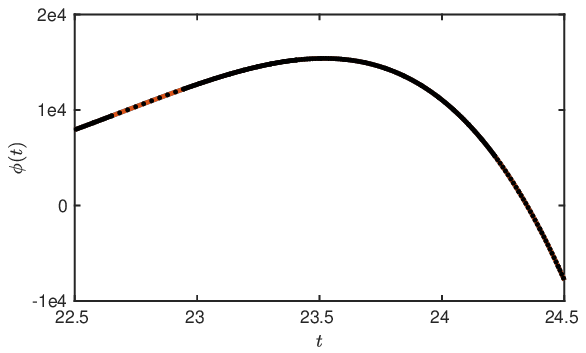}}
\caption{\label{FIGIVPexample2} {\bf IVP in Eq.\ \eqref{IVPexample2}} Comparison between (orange/solid) analytical solution and (black/dotted) ANDRe solution.}
\end{figure}

In contrast to the previous example, the IVP in Eq.\ \eqref{IVPexample2} is solved on an even larger domain with extensively increasing values. The results are shown in Fig.\ \ref{FIGIVPexample2} and aim to show that ANDRe is capable of solving time-integration problems on large domain with small neural networks. 
This is of particular importance as the domain size has been identified as an intricate parameter of the underlying problem, see also
the detailed study in \cite{Schneidereit2020Study}. \par
As displayed in Fig.\ \ref{FIGIVPexample2}, the ANDRe solution fits the analytical solution (orange) again on a qualitative and useful level. In total, the algorithm has finished after splitting the solution domain into 50 subdomains with the averaged $\ds \ell_1$-error of $\ds \Delta \phi_1=6.8152$e-4 (cf. Tab.\ \ref{TabNumResults}). In comparison, $\ds \Delta \phi_{\infty}$ differs more from $\ds \Delta \phi_1$ than the counterparts for IVPs in Eqs.\ \eqref{IVPexample1},\eqref{IVPexample3}. \par
While Fig.\ \ref{FIGIVPexample2a} shows the ANDRe solution for the entire domain, Figs.\ \ref{FIGIVPexample2b},\ref{FIGIVPexample2c} and \ref{FIGIVPexample2d} are zoomed in, to provide a more detailed view on certain areas. In Figs.\ \ref{FIGIVPexample2b} and \ref{FIGIVPexample2d}, we observe the local extreme points to be covered by more densely packed subdomains, especially the maximum in range of high function values. This may indicate, that the domain refinement not only depends on the complexity of a certain region, but on finding suitable minima in the weight space in order to get the training error below the verification error bound $\ds \sigma$. This however, does not hold for the extreme point in Fig.\ \ref{FIGIVPexample2c}. We see the local minimum to be covered by approximately equidistant subdomains (on a qualitative level) up to $\ds t=21.0362$. The next three subdomains however, are densely packed, only to be stretched again afterwards.

\begin{figure}[!h]
  \centering
  \subcaptionbox{\label{FIGIVPexample3a} entire domain}{
  \includegraphics[scale=1]{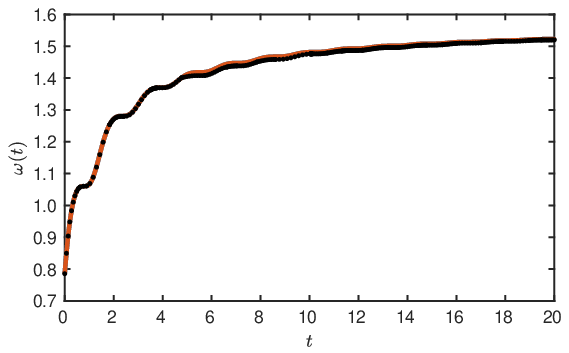}}
  \subcaptionbox{\label{FIGIVPexample3b} clipped domain}{
  \includegraphics[scale=1]{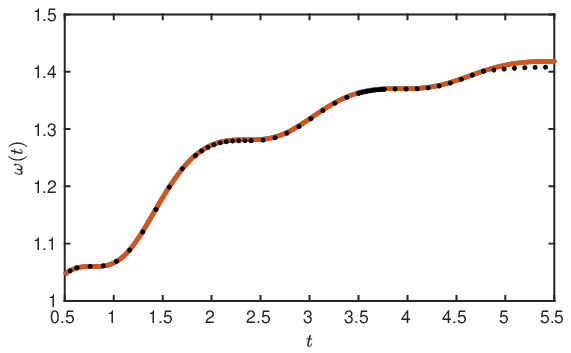}}
\caption{\label{FIGIVPexample3} {\bf IVP in Eq.\ \eqref{IVPexample3}} Comparison between (orange/solid) analytical solution and (black/dotted) ANDRe solution.}
\end{figure}

\begin{figure}[!h]
  \centering
  \subcaptionbox{\label{FIGIVPexample4a} Population over time, $\ds \tau(0)=3$, $\ds \kappa(0)=5$}{
  \includegraphics[scale=1]{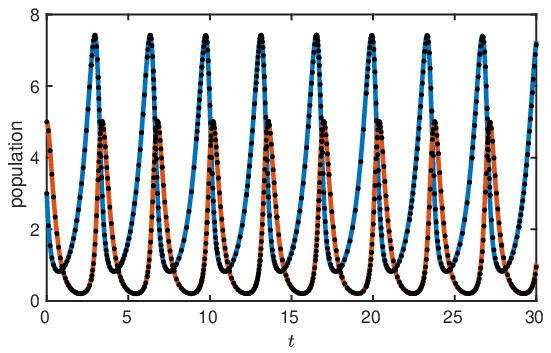}}
  \subcaptionbox{\label{FIGIVPexample4b} Comparison between $\ds \tau(t)$ and $\ds \kappa(t)$ in phase space with $\ds \tau(0)=3$ and (yellow) $\ds \kappa(0)=1$, (purple) $\ds \kappa(0)=3$, (green) $\ds \kappa(0)=5$.}{
  \includegraphics[scale=1]{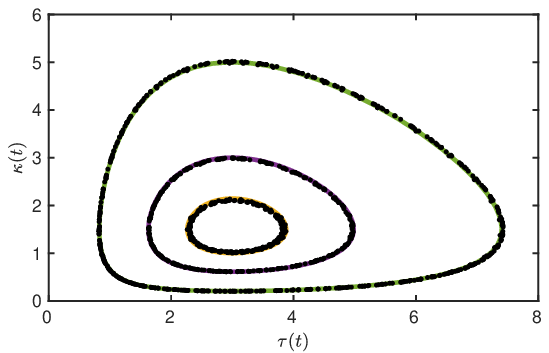}}
\caption{\label{FIGIVPexample4} {\bf IVP in Eq.\ \eqref{IVPexample4}} Comparison between (coloured/solid) Runge-Kutta 4 solution with 1e3 grid points and (black/dotted) ANDRe solution.}
\end{figure}

Results for the IVP in Eq.\ \eqref{IVPexample3} are displayed in Fig.\ \ref{FIGIVPexample3}. We decided to investigate this example because of the saddle points, which are repeatedly occurring. We observe a reliable solution approximation in the beginning of Fig.\ \eqref{FIGIVPexample3a}. However, from subdomain $\ds D_7$ and $\ds t=4.7760$ on, we can see that ANDRe starts to differ from the analytical solution. Although it keeps the general trend, and seems to converge against the analytical solution again in the end, the differences in this region are remarkable. This also marks a turning point computational-wise, which we will discuss more in detail in the corresponding experimental paragraph. Nonetheless, we had to decide to limit the verification error bound to $\ds \sigma=1e0$, since the computation with a lower error bound always got stuck around this area. This means both the Adam learning rate and the number of hidden layer neurons started to increase heavily. Although one would suggest, based on the universal approximation theorem, that at some point 
\begin{wrapfigure}[44]{r}{0.5\textwidth}  
\centering
  \subcaptionbox{\label{FIGerrIVPexample1} {\bf IVP in Eq.\ \eqref{IVPexample1}}}{
  \includegraphics[scale=0.95]{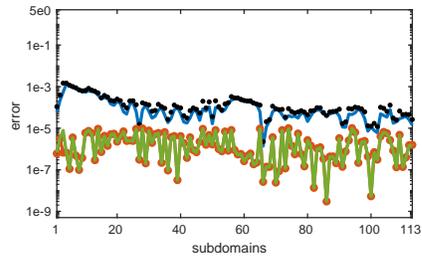}}
  \\
  \subcaptionbox{\label{FIGerrIVPexample2} {\bf IVP in Eq.\ \eqref{IVPexample2}}}{
  \includegraphics[scale=0.95]{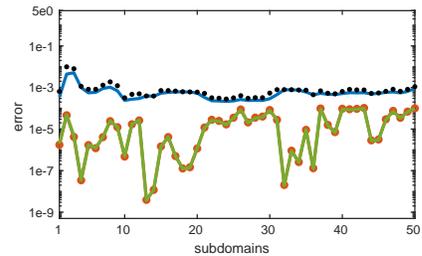}}
  \\
  \subcaptionbox{\label{FIGerrIVPexample3} {\bf IVP in Eq.\ \eqref{IVPexample3}}}{
  \includegraphics[scale=0.95]{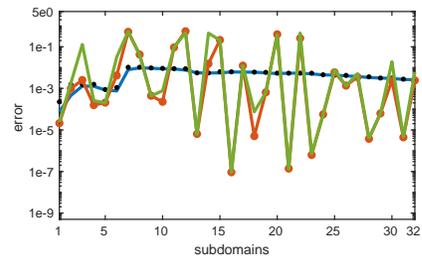}}
  \\
  \subcaptionbox{\label{FIGerrIVPexample4} {\bf IVP in Eq.\ \eqref{IVPexample4}}}{
  \includegraphics[scale=0.95]{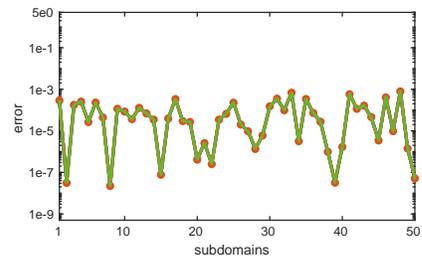}}
\caption{\label{FIGerrIVPexample} Error comparison, (blue/solid) numerical error, (black/dotted) infinity norm, (orange/marked) training error, (green/solid) verification error.}
\end{wrapfigure}
ANDRe would move on, we cancelled the time consuming computation at this point. This circumstance is definitely interesting to further investigate. We do not see a limitation of ANDRe here, since on a theoretical level, there should be an amount of hidden layer neurons, which is able to finish the computation even for a smaller verification error bound $\ds \sigma$. \par
Fig.\ \ref{FIGIVPexample3b} confirms the results from the previous examples, that the appearance of local extreme points (saddle points in this case) not necessarily result in more densely packed subdomains directly at their location. However, the final distribution has some packed subdomains inbound, prior to the local extreme points. The reason for this seems to be that a saddle point can not be part of a subdomain that is too large. Therefore the antecedent subdomain results in a smaller size so that the saddle point can be part of an appropriate sized subdomain. \par 
In Fig.\ \ref{FIGIVPexample4} both the Runge-Kutta 4 solutions and the ANDRe solutions are shown. For a fair comparison on the quantitative side, both method should use equal amount of training points, which in this case would arise from ANDRe solution. However, we are more interested in a qualitative comparison, since the Runge-Kutta 4 is known to provide very good results. ANDRe found a useful solution for the Lotka-Volterra equations in Fig.\ \ref{FIGIVPexample4a}, since there are only minor differences from the qualitative perspective. \par 
Fig.\ \ref{FIGIVPexample4b} shows the solution related to three different initial values for the predators. Let us note, that although  Fig.\ \ref{FIGIVPexample4b} only displays the solution at the training points (the same holds for the previous example IVPs), the trained SCNF is capable of evaluating the solution at every arbitrary discrete grid point over the entire domain, which is an advantage over numerical integration methods.

\paragraph{\bf Numerical and neural network errors.}

The measurement metrics (numerical and verification error) are highly relevant to discuss for ANDRe. In the subsequent diagrams we show the $\ds \ell_1$-error (blue/solid), the $\ds \ell_{\infty}$-error (black/dotted), as well as the verification error (green/solid) and the training error (orange/marked) over the successfully learned subdomains. \par 
Commenting on the relation between the verification and the training error in Fig.\ \ref{FIGerrIVPexample1} for the IVP in Eq.\ \eqref{IVPexample1} (cf. Eqs. \eqref{TPcost},\eqref{VPcost}), we see that both are mostly equal. This implies, that the corresponding subdomains have been effectively learned up to the desired state. Turning to the numerical errors, in regions where $\ds \Delta \psi_{1}$ shows an approximately constant slope, e.g., $\ds D_{58}$ to $\ds D_{64}$, $\ds \Delta \psi_{\infty}$ appears to deviate less from the averaged $\ds \ell_1$-error. Since the main goal of ANDRe is to make use of the verification error as a measurement metric for the numerical error, finding a relation between both is desirable. In Fig.\ \ref{FIGerrIVPexample1}, there are some regions that may indicate such a relation. The network errors from around $\ds D_{58}$ to $\ds D_{64}$ show a slightly decreasing behaviour, only to highly differ in $\ds D_{65}$. In comparison, $\ds \Delta \psi_{1}$ also slightly decreases for some subdomains and dips, together with the network errors, for $D_{65}$. However, in other regions almost no consistent relation is visible. \par
The statements made above also apply to Fig.\ \ref{FIGerrIVPexample2} for the IVP in Eq.\ \eqref{IVPexample2}, where one may find some relation in the beginning, while afterwards the $\ds \ell_1$-error is almost constant and the network errors drop and rise by several orders. \par
However, let us comment on the behaviour of both verification and training error, displayed in Fig.\ \ref{FIGerrIVPexample2}. While both match, they undergo the preset of $\ds \sigma$=1e-4 in some cases by several orders. Two adjacent subdomains may have 
\begin{wrapfigure}[17]{r}{0.5\textwidth}  
\centering
  \includegraphics[scale=1]{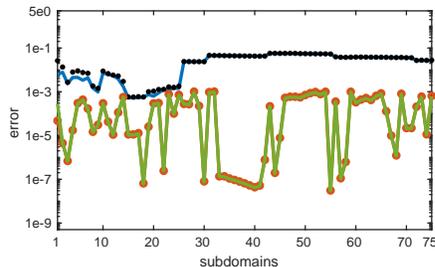}
\caption{\label{FIGerrIVPexample2_new} {\bf IVP in Eq.\ \eqref{IVPexample2}} Solved with ANDRe and alternative SCNF approach in Eq.\ \eqref{SCNF_mTSM} Error comparison, (blue) $\ds \ell_1$-error, (black/dotted) $\ds \ell_{\infty}$-error, (orange/marked) training error, (green/solid) verification error.}
\end{wrapfigure}
a verification/training error with significant differences. Although we find a local maximum for the $\ds \ell_1$-error in the beginning, it decreases afterwards and remains in a certain region with local minima and maxima. However, we find the decreasing behaviour to be an important characteristic compared to numerical methods, where one would expect the error to accumulate. \par
Commenting on Fig.\ \ref{FIGerrIVPexample3} for the IVP in Eq.\ \eqref{IVPexample3}, the possible local relations between the numerical and network errors seem to have turned into a chaotic state. For the corresponding verification error bound $\ds \sigma=1e0$, even the network errors are most of the time not equal anymore. In contrast the both Fig. \ref{FIGerrIVPexample1} and Fig.\ \ref{FIGerrIVPexample2}, the numerical errors lay in between the values of the network errors, which is highly interesting. This more or less confirms, that even if the verification/training error indicate a shallow (local) minimum in the weight space, the numerical error can still be useful. Although $\ds \Delta \omega_1$ and $\ds \Delta \omega_{\infty}$ are, expect for the beginning, almost constant throughout the domain. That circumstance inherents both good and bad news. The latter connects to the apparent random behaviour, while the good news is that even though the network errors appear to be random, the IVP can still be considered to be solved. \par 
The results for the Lotka-Volterra equations in Fig.\ \ref{FIGerrIVPexample4} also seem to indicate a chaotic behaviour. That is, the local minima of orders around $\ds \approx$1e-8 relate to arbitrary subdomains, that are not connected to, e.g., the periodic extreme points of the solution. \par   
The diagram in Fig.\ \ref{FIGerrIVPexample2_new} shows the results for a different SCNF approach \cite{piscopo2019solving,Schneidereit2021SCNF}, combined with ANDRe. In contrast to the neural forms approach described in Section \ref{sectionSCNF} (using the initial condition to construct the neural form), now we directly combine with neural networks with the polynomial ansatz \cite{Schneidereit2021SCNF,Schneidereit2020Study}:
\begin{equation}
\tilde{\phi}_C(t_{i,l},\mat{P}_{m,l})=N_1(t_{i,l},\vecp_{1,l})+\sum_{k=2}^mN_k(t_{i,l},\vecp_{k,l})(t_{i,l}-t_{0,l})^{k-1}
\label{SCNF_mTSM}
\end{equation}
Since the initial condition is not included in Eq.\ \eqref{SCNF_mTSM}, it appears as an additional term directly in the cost function. Here, we use 
\begin{equation}
g(t)=\frac{\ds 1+\frac{1}{1000}e^t\cos(t)}{1+t^2}
\end{equation}
as a shortcut for:
\begin{equation}
\begin{array}{cc}
\ds E_l[\mat{P}_{m,l}]=\frac{1}{2(n+1)}\sum_{i=0}^n&\ds \bigg\{ \dot{\tilde{\phi}}_C(t_{i,l},\mat{P}_{m,l})+g(t)+\frac{2t}{1+t^2}\tilde{\phi}_C(t_{i,l},\mat{P}_{m,l})\bigg\}^2+ \\  
\ds &\hspace*{-2.55cm}\ds\frac{1}{2} \bigg\{ N_1(t_{0,l},\vecp_{1,l})-\tilde{\phi}_C(t_{0,l},\mat{P}_{m,l})\bigg\}^2 
\end{array}
\end{equation}
Hence, the initial condition is learning by the first neural network. The cost function construction concept is very similar to physics-informed neural networks \cite{Raissi2019PINN,Jagtap2020cPINN}. However, the polynomial approach ansatz in different in this context. Let us recall, that the initial values follow $\ds \tilde{\phi}_C(t_{0,1},\mat{P}_{m,1})=\phi(0)$ in the leftmost subdomain and $\ds \tilde{\phi}_C(t_{0,l},\mat{P}_{m,l})=\tilde{\phi}_C(t_{n,l-1},\mat{P}_{m,l-1})$ elsewhere. This concept avoids possible difficulties in constructing a suitable neural form. Since both approaches (Eq.\ \eqref{SCNF} and Eq.\ \eqref{SCNF_mTSM}) only differ in their cost function construction, it appears natural to compare them. That is, the results in Fig.\ \ref{FIGerrIVPexample2_new} compare to Fig.\ \ref{FIGerrIVPexample2}. The behaviour of both verification/training error does not seem to be connected to the numerical error, for both methods. Fig.\ \ref{FIGerrIVPexample2_new} shows less accurate results for the $\ds \ell_1$-error. The loss of accuracy possibly relates to the fact, that here the new initial condition for the next subdomain is not fixed by adding it to the neural form. It rather has to be learned again, which in practice may harm the usefulness of this approach. The gap between both $\ds \ell_1$-error and $\ds \ell_{\infty}$-error closes at a certain point. \par
\begin{wrapfigure}[18]{r}{0.5\textwidth}  
  \centering
  \includegraphics[scale=1]{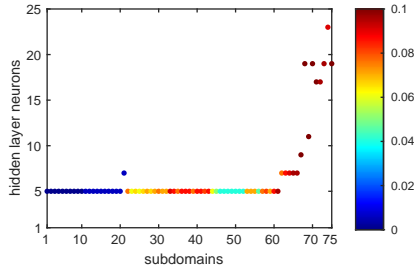}
\caption{\label{FIGneuronsIVPexample2_new} {\bf IVP in Eq.\ \eqref{IVPexample2}} Solved with ANDRe and alternative SCNF approach in Eq.\ \eqref{SCNF_mTSM} Visualisation of the automatic parameter adjustment (hidden layer neurons and learning rate) over the subdomains, (coloured) Adam learning rate $\ds \alpha$.}
\end{wrapfigure}
When turning to Fig.\ \ref{FIGneuronsIVPexample2_new}, we observe that the parameter adjustment  brought the Adam learning rate (coloured) up to various values in order to finish learning the subdomains. Additionally, the necessary number of hidden layer neurons also heavily increases towards higher subdomains. \par 
Although the results in terms of the numerical error are not better than in Fig.\ \ref{FIGerrIVPexample2}, we find here a confirmation of the automatic parameter adjustment. With this feature, ANDRe was able to solve the IVP. That is, we see our approach to enable the parameter adjustment when necessary, to be justified by the results. However, this does not support the overall usage of this alternative SCNF approach in context of ANDRe.

\paragraph{\bf Method and parameter evaluation.}

In this paragraph we investigate and evaluate different parts of the method. \par

\begin{wrapfigure}[13]{r}{0.5\textwidth} 
  \centering
  \includegraphics[scale=1]{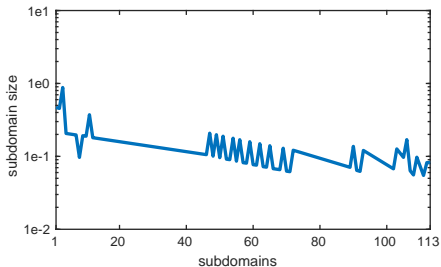}
\caption{\label{FIGsubsizeIVPexample1} {\bf IVP in Eq.\ \eqref{IVPexample1}} Visualisation of the learned subdomain sizes.}
\end{wrapfigure}

In Fig.\ \ref{FIGsubsizeIVPexample1}, the sizes of the learned subdomains for the IVP in Eq.\ \eqref{IVPexample1} are shown. The general trend points towards smaller subdomains throughout the computation. However, we witness that there are local differences with bigger or smaller subdomains and this is what we expect from ANDRe. The subdomain size is reduced until it is sufficiently small and that can be individual for each part of the solution. Nonetheless, let us compare both the numerical error in Fig.\ \ref{FIGerrIVPexample1} and the subdomain sizes in Fig.\ \ref{FIGsubsizeIVPexample1}. In the first ten subdomains there seems to be a certain correlation, a larger size in this range results in a larger numerical error. A smaller verification error bound $\ds \sigma$ to deal with the discrepancy between verification and training error in Fig.\ \ref{FIGerrIVPexample1} may have resulted in another size reduction with better results. However, the statement that a smaller (local) subdomains size implies a better numerical error does not hold here. Although one of the complex conditions employed each subdomain to not become smaller than 0.1, we can see in Fig.\ \ref{FIGsubsizeIVPexample1} that certain subdomains towards the end undergo this preset condition. This is related to the fact, that the subdomain size is first reduced and then checked for its size. Therefore it is still possible for a subdomain, to slightly undergo the size of 0.1. However, and as the results confirm, a further size reduction is not possible.

\begin{wrapfigure}[15]{r}{0.49\textwidth}
  \centering
  \includegraphics{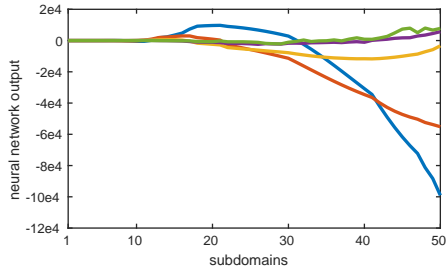}  
  \caption{\label{FIGnetoutIVPexample2} {\bf IVP in Eq.\ \eqref{IVPexample2}} Output of the incorporated SCNF neural networks, (blue) $\ds N_1$, (orange) $\ds N_2$, (yellow) $\ds N_3$, (purple) $\ds N_4$, (green) $\ds N_5$.}
\end{wrapfigure}
Turning to Fig.\ \ref{FIGnetoutIVPexample2}, the (learned) neural network outputs are displayed for the incorporated SCNF (cf. Eq.\ \eqref{SCNF}) order $\ds m=5$ of the IVP in Eq.\ \eqref{IVPexample2} (cf. Fig.\ \ref{FIGIVPexample2}). That is, the five displayed graphs each represent one neural network output over the subdomains. We see the first and second SCNF orders to dominate the results for higher numbers of subdomains. However, higher orders also contribute to the solution, making our approach adaptive in the approximation order, indirectly. Let us recall, that the factors $\ds (t_{i,l}-t_{0,l})^k$ for the different neural networks $\ds N_k(t_{i,l},\vecp_k), k=1,\ldots,5$, dictate the impact of each neural network since they act as a scaling factor. Hence, subdomains with a size below 1 imply a smaller influence of higher SCNF orders. This circumstance can be challenging for an IVP solution with large values. Nonetheless, we see that our SCNF algorithm was able to solve the IVP in Eq.\ \eqref{IVPexample2}, even tough it incorporates fairly large values. \par

\begin{table}[!h]
\centering
\caption{{\bf IVP in Eq.\ \eqref{IVPexample1}} Results for a complete learning procedure for one subdomain.}
\begin{tabular}{c|c|c|c|c|c}
$\ds t_{25}$ & $t_{26}$ & $\ds \Delta \psi_1$ & $\ds E_{25}^{TP}$ & $\ds E_{25}^{VP}$ & $\ds \alpha$ \\
\hline
5.5953 & 15.000 & 1.3557 & 22.888 & 23.598 & 1e-3 \\    
5.5953 & 10.298 & 2.3780 & 1.7622 & 22.695 & 1e-3 \\
5.5953 & 7.9465 & 1.3774 & 6.8529 & 34.448 & 1e-3 \\
5.5953 & 7.9465 & 0.3868 & 10.640 & 9.9730 & 6e-3 \\
5.5953 & 6.7709 & 3.4447e-2 & 3.2209e-2 & 3.9242e-2 & 6e-3 \\
5.5953 & 6.1831 & 1.1881e-2 & 3.4791e-2 & 3.5682e-2 & 6e-3 \\ 
5.5953 & 5.8892 & 3.3488e-4 & 1.0875e-4 & 1.0779e-4 & 6e-3 \\
5.5953 & 5.7423 & 1.6882e-4 & 2.4565e-6 & 2.3747e-6 & 6e-3 \\
\end{tabular}
\label{TABOneSubdomain}
\end{table}

In Tab.\ \ref{TABOneSubdomain}, quantitative results for the entire learning process of one subdomain of the IVP in Eq.\ \eqref{IVPexample1} are displayed. The left subdomain boundary $\ds t_{25}$ remains constant while the right subdomain boundary $\ds t_{26}$ is adjusted as in Eq.\ \eqref{adjustBoundary}. The verification error values $\ds E_{25}^{VP}$ demonstrate the appearance of non-uniform learning during the solution process and show how important the verification error and the parameter adjustment are. While $\ds E_{25}^{TP}$ decreases (as intended) for the first two subdomain size reductions, it increases for the third one, which leads to a growth of the initial learning rate $\ds \alpha$. Now for the same subdomain size, $\ds E_{25}^{VP}$ decreased significantly (while $\ds _{25}^{TP}$ has increased again). That circumstance enables ANDRe to continue reducing the subdomain size until it is sufficiently small. \par 

\begin{table}[!h]
\centering
\caption{\label{TABDiffSigma} {\bf IVP in Eq.\ \eqref{IVPexample2}} Results for different $\ds \sigma$, on domain $\ds t\in[0,25]$.}
\begin{tabular}{c|c|c|c|c}
$\ds \sigma$ & $\ds h$ &$\Delta \phi_1$ & $\ds E^{VP}[\mat{P}_{m,l}]$ & $\ds E^{TP}[\mat{P}_{m,l}]$  \\
\hline
1e-1 & 37 & 5.5512e-2 & 1.8907e-2 & 1.8537e-2 \\
1e-2 & 39 & 1.0749e-2 & 1.5714e-3 & 1.6255e-3 \\
1e-3 & 47 & 6.2122e-3 & 1.1582e-4 & 1.1917e-4 \\
1e-4 & 50 & 6.8152e-4 & 2.6581e-5 & 2.7788e-5 \\
1e-5 & 59 & 3.0005e-4 & 1.9260e-6 & 2.2057e-6 \\
1e-6 & 74 & 1.1870e-4 & 2.1157e-7 & 2.2076e-7 \\
\end{tabular}
\end{table}

From the perspective of employing a condition for minimising the cost function, the question arises how the algorithm outcome is affected by different error bound values $\ds \sigma$. Tab.\ \ref{TABDiffSigma} shows the overall $\ds \ell_1$-error, verification error and training error for different $\ds \sigma$ regarding the IVP in Eq.\ \eqref{IVPexample2}. The choice of $\ds \sigma$ has a direct impact on each error value, as they all decrease the smaller $\ds \sigma$ gets. However, an experiment for $\ds \sigma=1e$-7 did not finish learning the subdomains. We terminated the computation after the number of hidden layer neurons crossed fifty one. In this subdomain, the smallest verification error was 1.9881e-7 but the optimisation did not manage to go below $\ds \sigma=1$e-7. This phenomenon may again relate to the complexity of the cost function energy landscape. Either such a local minimum could not be found by the optimiser for various reasons, or even the global minimum is still too shallow for that error bound. \par
Results in Tab.\ \ref{TABDiffSigma} confirm the results from \cite{Schneidereit2021SCNF}, where an increasing number of subdomains shows a decreasing numerical error. \par

\begin{wraptable}[17]{r}{0.49\textwidth}
\centering
\caption{\label{TABdeltaValues} {\bf IVP in Eq.\ \eqref{IVPexample3}} Results for different domain size reduction parameter values $\ds \delta$, on domain $\ds t\in[0,5]$.}
\begin{tabular}{c|c|c}
$\ds \delta$ & $\ds h$ & $\Delta \omega_1$  \\
\hline
0.9 & 5 & 6.2398e-3  \\
0.8 & 5 & 7.0250e-4  \\
0.7 & 4 & 3.4629e-3  \\
0.6 & 5 & 9.1571e-2  \\
0.5 & 4 & 6.9541e-4  \\
0.4 & 5 & 6.2569e-3  \\
0.3 & 5 & 4.4258e-3  \\
0.2 & 7 & 1.1200e-3  \\
0.1 & 13 & 2.7217e-4 \\
\end{tabular}
\end{wraptable}

Last but not least we discuss experimental results for different domain resize parameter values $\ds \delta$ in Tab.\ \ref{TABdeltaValues} for the IVP in Eq.\ \eqref{IVPexample3}. The higher this value is set, the more aggressive each subdomain is reduced in size. On the other Hand, the smaller $\ds \delta$ is, the more careful the subdomains are reduced in size. However, one would expect the necessary amount of subdomains to increase, the higher the resize parameter is. But in reality the results and the amount of subdomains are comparable for all $\ds \delta$, if we exclude $\ds \delta=0.1$. On the $\ds \ell_1$-error side, except for $\ds \delta=0.6$, all the results are comparable. Although the results are highly problem specific and may change with a larger domain size, we find $\ds \delta=0.5$ to provide the best mix with $\ds h=4$ and $\ds \Delta \omega_1=6.9541$e-4. This domain size parameter was used for all the computations.

\section{Conclusion and future work}

The proposed ANDRe is based on two components. First, the resulting verification error arising from the inbound subdomain collocation neural form (SCNF) acts as a measurement metric and refinement indicator. The second component is the proposed algorithm which refines the solution domain in an adaptive way. We find ANDRe to be a dynamic framework adapting the complexity of a given problem. In this paper, we have shown that the approach is capable of solving time-dependent differential equations of different types, incorporating various interesting characteristics, in particular including large domains and extensive variations of solution values. \par  
In contrast to numerical solution methods for solving initial value problems, the numerical error does not inevitable accumulate over the subdomains. It can rather decrease again due to the flexibility of the neural forms approach. A significant advantage of ANDRe is the verification step to make sure that the solution is also useful outside of the chosen training points. All this makes ANDRe a unique and conceptually useful framework. \par
However, several questions remain open for future work. While there seems to be a certain and natural correlation between the neural network and the numerical error, in reality this correlation appears to be sometimes a sensitive issue. It is unclear yet, whether some minima in the cost function energy landscape contribute better to the numerical error, or not. However, we find the verification error to already serve as a useful error indicator in ADNRe. In addition, we would like the numerical error to proportionally correspond to the neural network verification. If we could manage to achieve an improvement in the correlation between both errors or understand the relation more in detail on the theoretical level, we think that the ANDRe approach can perform even better in the future. 
\par
We also find relevant to further investigate the computational parameters and fine tuning the parameter adjustment part of ANDRe. The verification step may be considered as a part in the optimisation process, to predict early, whether a further optimisation in the corresponding subdomain is useful or a size reducing is mandatory. This could lower the computational cost but has to be incorporated and tested carefully to not lose any information during the optimisation process. \par 
Since ANDRe represents an additional discretisation in time, the approach should also work for PDEs with both time and spatial components and it appears natural to extend in future work the method to multidimensional differential equations.

\section*{Acknowledgement}
This publication was funded by the Graduate Research School (GRS) of the Brandenburg University of Technology Cottbus-Senftenberg. This work is part of the Research Cluster Cognitive Dependable Cyber Physical Systems.

\end{document}